# Artificial Intelligence-Based Analysis of Ice Cream Melting Behavior Under Various Ingredients

Zhang Lai Bin[1] and Zhen Bin It[2*]

[1] PSB Academy, Singapore
[2] Singapore University and Technology and Design, Singapore
[2] zhenbin_it@mymail.sutd.edu.sg

**ABSTRACT**

The stability of ice cream during melting is a critical factor for consumer's acceptance and product quality. With the commonly added stabilizer to improve texture, structure and slower melting as the factors to analyze. This report explores the effects of locust bean gum, guar gum, maltodextrin, and carrageenan on the melting behavior of homemade ice cream. The main objective was to assess how these additives influence melting resistance and to identify a more cost-effective recipe formulation. Ice cream samples incorporating each additive were prepared and subjected to melting tests under controlled conditions. Timelapse recordings were used to capture and analyze the progression of melting over time. Python and OpenCV is used for process and analysis.

Observations revealed that all samples retained a foam-like structure even after melting, suggesting the stabilizers contributed to the formation of a stable air-cell matrix. Furthermore, when the melted samples were re-frozen and subsequently melted again, they displayed increased sturdiness, indicating improved resilience of the ice cream structure. Comparative analysis of the different stabilizers highlighted variations in their effectiveness, with some offering stronger melting resistance and structural support than others.

Overall, the findings provide insights into the functional roles of commonly used food additives in ice cream formulation. By evaluating both performance and cost, this study demonstrates the potential for developing recipes that balance durability with economic efficiency, contributing to practical applications in both small-scale and commercial ice cream production.

## Introduction

This study investigates the melting behavior of ice cream formulations with varying stabilizing ingredients and temperature conditions, incorporating computer vision techniques for objective analysis. The primary objective is to evaluate how different additives—such as guar gum, carrageenan, locust bean gum, maltodextrin and basic ingredients—affect structural stability and melting resistance. Ice cream samples were prepared with controlled compositions and subjected to melting tests at ambient temperatures ranging from 25°C to 35°C. The Convolutional Neutral Network (CNN) was trained to monitor the melting process through time-lapse imaging and python programming analysis. Parameters such as melted surface area, drip flow rate, and melting progression over time were extracted using contour detection and pixel-based measurement techniques. The results demonstrate that stabilizers significantly influence melting behavior, with gum-based formulations exhibiting superior thermal resistance. The computer vision approach provided consistent, real-time, and non-invasive monitoring, enabling precise comparison between samples. These findings offer a practical method for optimizing ice cream formulations for enhanced performance in warm environments and extended shelf life during storage and transport

## Literature Review

The following section will investigate the recent studies and findings to conduct experiment in the melting of ice cream and training of models to aid in the analysis of this project.

### Fat Crystallization

In one of the recent studies highlights the fat composition with the help of emulsifiers and crystallization behaviour, which determine the melting resistance of ice cream. One of the main reasons for ice cream to be soft, smooth texture and melting resistance is the partial-fused fat globules that was crystalized forms a network structure that are critical in holding the shape of the ice cream and obtaining desirable melting properties [1]. With the acknowledgement of the fat globule's role in the process of making ice cream, it is important to learn the details. During the whipping process, air is trapped and locked into the aggregated fat globules, forming strong network structures and stabilize as the whipping process continues. Moreover, the fat crystal in this process is also rupturing the interfacial membrane of the fat globules creates the aggregated fat globules that plays a crucial role in balancing the air-water interface [1-2]. Fat structure and mix viscosity play a central role in determining creaminess and melt resistance [4]. In this report, the main fat ingredient will be the basic ingredient in making the ice cream, they consist of the egg yolk, milk and cream. Forming up the high percentage of crystalline fat, creating the base structure of ice cream in this project.

**Types of Ice Cream Additives**

Apart from the basic ingredients, many manufactures will add additives and flavouring to make the ice cream more delicious and extent shelve life. To look into the wide range of ingredients and come up with different combination of ice cream can be fascinating. For instance, the fat sources can be explored finding an alternative replacement for milk cream. Vegan ice cream can use extra virgin olive oil to create a "Mediterranean ice cream" [10]. Wide choices of oil show how fat substitution can alter both the nutritional and physical melting properties. Stabilizers and emulsifiers can further enhance the structural stability of ice cream, developing optimized mixtures to improve viscosity, and melting resistance [20]. The use of functional and probiotic ingredients, such as mare's milk in yogurt ice cream [3] or bifidobacteria with prebiotics [8], introduces additional complexity by altering properties that may influence the melting of ice cream. With the inspiration of yogurt ice cream, this project will create a comparing test set consisting of yogurt for analysis.

Proteins and hydrocolloids also play a crucial role in determining the texture of ice cream and melt resistance, as the stabilize air cells and water phases within the frozen matrix. Studies on alternative protein sources, such as Phaseolus vulgaris flour, have shown improved foaming properties and melting stability when used in ice cream formulations [11]. Hydrocolloids including guar gum, carrageenan, and locust bean gum are widely used to control viscosity and inhibit ice recrystallization, reducing rapid meltdown during storage and consumption. highlighted how protein–polysaccharide interactions enhance emulsion stability [9], while optimized stabilizer mixtures helps to improve overrun and melting resistance [20]. These findings emphasize that the stabilizers and proteins not only provide structural reinforcement but also influence consumer perception by slowing meltdown, maintaining creaminess, and extending shelf life.

**Machine learning Algorithms**

Machine Learning (ML) is a branch of artificial intelligence that enables systems to learn from data and make decisions without explicit programming. Melting behaviour is affected by many factors such as heat transfer, ice recrystallization, and phase transitions during storage and consumption. Some research highlighted the physical principles of ice cream melting, linking the compositional changes to thermal responses [7]. Recent advanced analysis techniques, such as calorimetry, rheological measurements, and microscopic imaging, have improved understanding of ice cream's dynamic behaviour under different temperatures [13,15]. With the rise of artificial intelligence, predictive modelling and image-based analysis are promising tools for faster and more accurate evaluations in comparison to the traditional manual methods. Such AI-based analysis approaches can be applied to food science fields, enabling automated monitoring of ice cream meltdown under various ingredients and storage conditions.

**Economic considerations**

Beyond technical and sensory qualities, the economics of ice cream production play a crucial role in determining ingredient selection and product viability. Food additives and stabilizers such as guar gum, carrageenan, and locust bean gum are widely used and well excepted. These stabilizers not only help in controlling meltdown but also a more cost-effective option compared to more expensive fat or protein additions [20]. Alternative fat sources, such as replacing milk cream with extra virgin olive oil, have been studied for their nutritional benefits, consumers from wealthier class are more likely to purchase these healthier products. However, the cost implications remain a challenge for large-scale application and the production cost and profit may not be desirable. [10]. Similarly, the incorporation of unconventional dairy sources like mare's milk [3], goats milk or functional cultures such as probiotics [8], unique fruit choice such as dragon fruit can enhance nutritional and market appeal [22], but these may increase the raw material costs transportation fees and production complexity. Balancing the costs with consumer's demand for health, functionality, taste, texture and melting stability are essential in sustainable ice cream manufacturing, particularly as the market becomes more competitive and ingredient prices fluctuates.

### AI approach

To understand the how ice cream melt, long observations are required, and this process is not feasible to observe by human eyes. Hence, with the advancement of artificial intelligence (AI) model development, high image processing performance and analysis, it is the suitable tool. The increasing use of AI [23] and sensor systems in engineering and biomedical applications shows that it has strong potential for adoption in food analysis, including the ice cream melting studies [24]. Recent studies highlight the importance of development in AI models for image classification, which is directly relevant when applying computer vision techniques to monitor the melting patterns. Similarly, exploring the AI-based soft sensing in non-Newtonian fluid systems, shows how sensor integration with machine learning can capture the complex flow behaviours [25]. This can be an approach that could be adapted for the rheological properties of ice cream during melting. Furthermore, sensor innovations such as the low-cost capacitance sensor for bubble monitoring in fluids [27], also suggest the possibilities for real-time measurement on the structural changes in frozen desserts. Moreover, advancements of transfer learning and deep neural networks for signal detection [28] illustrates how robust AI frameworks can enhance the reliability for cross-sample predictions, supporting the development of models for ice cream melting analysis. Another aspect which is important will be the implementation of power and heat management systems [29-30] which will help in the thermal management of how the ice cream will melt over time.

### Our proposed study

This research paper propose aims to analyse the melting behaviour of ice cream under various ingredient formulations by introducing both experimental and computational approach. Different stabilizers and ingredient will be added to the ice cream formulations, and their effects on the physical stability and melt resistance will be evaluated. Real-time monitoring methods and image-based analysis will be employed to track the melting process. While the artificial intelligence models will be applied to identify the melting behaviour under different formulation. This study is expected to provide insights into the ingredient interactions on ice cream stability and establish an AI-driven methods tools for food quality evaluation.

### Experiment ingredients and procedure

#### Bill of materials

Bill of materials including the waste samples that are not analysed:

Table 1 Bills Of Material For Ice Cream

| Items | Description | Quantity Used | Unit Price | Total Cost For Used Quantity |
|---|---|---|---|---|
| 1 | Meiji Full Cream Milk | 250 ml | $6.70/ 2 litres | $0.84 |
| 2 | Pure Cane Sugar | 360 g | $4.89 / 3 kg | $0.59 |
| 3 | Whipping Cream | 600 ml | $4.35/ 200 ml | $13.05 |
| 4 | Egg yolk | 9 eggs | $6.65/ 30 eggs | $2.00 |
| 5 | Strawberry Flavoured Yogurt containing oats | 150 g | $12.90/ 2000 g | $0.97 |
| 6 | Guar Gum | ~5g | $0.93 /100g | $0.05 |
| 7 | Carrageenan | ~5g | $2.16 /50g | $0.22 |
| 8 | Locust bean gum | ~2g | $1.60 /20g | $0.16 |
| 9 | Maltodextrin | ~2g | $0.52 /20g | $0.05 |

From the table above, apart from the crucial ingredients such as milk, cream, sugar and eggs, the selection of additives and extra ingredient can be selected just based on its price. The Guar Gum and Maltodextrin is a much cheaper alternative than compared to Carrageenan and Locust Bean Gum. With the recorded analysis, the adding of 2 additives make the structure of ice cream strong and sturdy even after melting. Hence with the combinations out of all 4 additives, Guar Gum and Maltodextrin is the cheaper combination that works well at slowing the melting of ice cream and maintaining the structure of the ice cream.

#### Procedure for plain ice cream
1. Break 3 eggs and separate the egg yolks
2. Add 60 g sugar to the egg yolks and stir the mixture until it becomes slightly lighter in colour
3. Warm 60 g milk in a pot until small bubbles appear at the edges

4. Slowly add the hot milk into the egg mixture and stirring the mixture constantly
5. Pour the mixture back into the pot and heat gently, keep stirring until slightly thicken
6. Remove from heat and allow the mixture to cool at the side
7. In a clean bowl, pour 250 g cream and add 10 g sugar to the cream
8. Whip the cream until firm peaks form
9. Gently pour the cooled egg mixture into the whipped cream until it is smooth and well blended
10. Take out small samples of the base mixture and add different additives and combination to each sample
11. Pour them into mould and freeze the mixtures

**Procedure for yogurt ice cream**
1. Break 3 eggs and separate the egg yolks
2. Add 50 g sugar to the egg yolks and stir the mixture until it becomes slightly lighter in colour
3. Warm 80 g milk in a pot until small bubbles appear at the edges
4. Slowly add the hot milk into the egg mixture and stirring the mixture constantly
5. Pour the mixture back into the pot and heat gently, keep stirring until slightly thicken
6. Remove from heat and allow the mixture to cool at the side
7. In a clean bowl, pour 300 g cream and add 45 g sugar to the cream
8. Whip the cream until firm peaks form
9. Gently pour the cooled egg mixture and 150g of yogurt into the whipped cream until it is smooth and well blended
10. Take out small samples of the base mixture and add different additives and combination to each sample
11. Pour them into mould and freeze the mixtures

Table 2 Recipe for ice cream

| Ingredients | Set 1 | Set 2 |
|---|---|---|
| Milk | 60 g | 80 g |
| Sugar | 70 g | 95 g |
| Cream | 250 g | 300 g |
| Egg yolk | 3 | 3 |
| Strawberry Flavoured Yogurt containing oats | - | 150 g |

The ingredients are added according to the ratio, yogurt ice cream as a flavouring dilutes the fat content ratio, hence more cream and milk and sugar are added to allow good structure of ice cream to be formed. The strawberry flavoured yogurt containing oats was a carefully made choice, the yogurt can provide strawberry flavours to the ice cream and the oats inside the yogurt helps to provide a firm structure support for the ice cream.

## Preprocessing

To evaluate the effects of different stabilizers and additives on ice cream melting resistance, a series of controlled formulations were prepared. Two sets of samples were designed for data collection, Set 1 (Plain Ice Cream) and Set 2 (Yogurt Ice Cream). Each set consisted of nine samples with varying combinations of carrageenan, guar gum, locust bean gum, and maltodextrin. These additives were selected because of their well-known performance in improving viscosity, stabilizing emulsions, and reducing ice recrystallization in frozen desserts.

Table 3 Experiment Set Table

| Samples | Set 1 (plain) | | | | Set 2 (Yogurt) | | | |
|---|---|---|---|---|---|---|---|---|
| | Carragean (0.01-0.3% by weight) | Guar (0.1-0.3% by weight) | Locust (0.1-0.3% by weight) | Maltodextrin (2-10% by weight) | Carragean (0.01-0.3% by weight) | Guar (0.1-0.3% by weight) | Locust (0.1-0.3% by weight) | Maltodextrin (2-10% by weight) |
| 1 | - | - | - | - | - | - | - | - |
| 2 | ✓ | - | - | - | ✓ | - | - | - |

| 3 | - | ✓ | - | - | - | ✓ | - | - |
| 4 | ✓ | ✓ | - | - | ✓ | ✓ | - | - |
| 5 | ✓ | - | ✓ | - | ✓ | - | ✓ | - |
| 6 | ✓ | - | - | ✓ | ✓ | - | - | ✓ |
| 7 | - | ✓ | ✓ | - | - | ✓ | ✓ | - |
| 8 | - | ✓ | - | ✓ | - | ✓ | - | ✓ |
| 9 | ✓ | ✓ | ✓ | ✓ | ✓ | ✓ | ✓ | ✓ |

To minimize the material costs and simplify the experiment, most formulations were limited to a maximum of two additive combinations, except for Sample 9, which contained all four additives to provide a reference and comparison for full stabilization.

## Data Collection

To ensure consistency and accuracy of the data collected from the ice cream samples. Each formulation (plain and yogurt-based) was first standardized in terms of preparation, sample mass, and storage conditions. All samples were stored in the same mould at the same freezer temperature.

For the melting experiments, each sample was recorded by specific time measurement (30 min). Images of the melted ice cream were also captured to provide additional visual data. These raw measurements were then organized into structured datasets in the PyCharm. Furthermore, to reduce human errors and ensuring fair comparison, all the recording was done at fix time range to avoid the influence of weather temperature.

## Proposed machine learning algorithm

This project utilizes a free and open-source software, PyCharm. To develop a machine learning-based model that helps to extract specific frame images from timelapse video and trained to categorize the following images into its specific state, such as solid, partial and melted. The implementation is informed and inspired by existing research studies conducted by other scholars in the field. With the relevant research studies, simple training of the CNN model is feasible and allow the analysis of huge amount of recorded data to be much more efficient. With the sorted data, the melting time of the samples can be viewed clearly.

## Training and Validation

In this study, Python and CNN was applied to classify the physical state of ice cream under different ingredient formulations. The dataset, prepared from plain and yogurt-based ice cream samples with varying stabilizer combinations, was divided into two subsets and used to train the CNN. The train set consisted half of the plain ice cream data, which was feed into the solid, partial and melted state folder. With the base image, the CNN model can be trained to process and divide the rest of the data sets accordingly.

### Initial testing stage

In this stage, many steps and experiments are unclear and not precise. The initial stage of the project was full of problems. With low contents of fat globules, the water content does not mix well with the other ingredients. Moreover, with the lack of whisk or ice cream machine it is clearly shown in the picture below that the mixture is not well mixed and they separate into layers during freezing. Furthermore, the mould is too hard to take out the samples and the samples are too large for experiments. The melting of large mould sample could take up to 1 hour to melt completely. During the recording of the process, the camera stops recording after every 20 min due to unknown reasons, causing interruption of the recording and multiple recordings are taken for just one sample. Storage of videos is a problem as the recording is taking up a lot of space in both the camera and my computer. Hence, consideration of recording timelapses with smaller quantity of ice cream to reduce time and storage is preferred, and it will be easier for analysis.

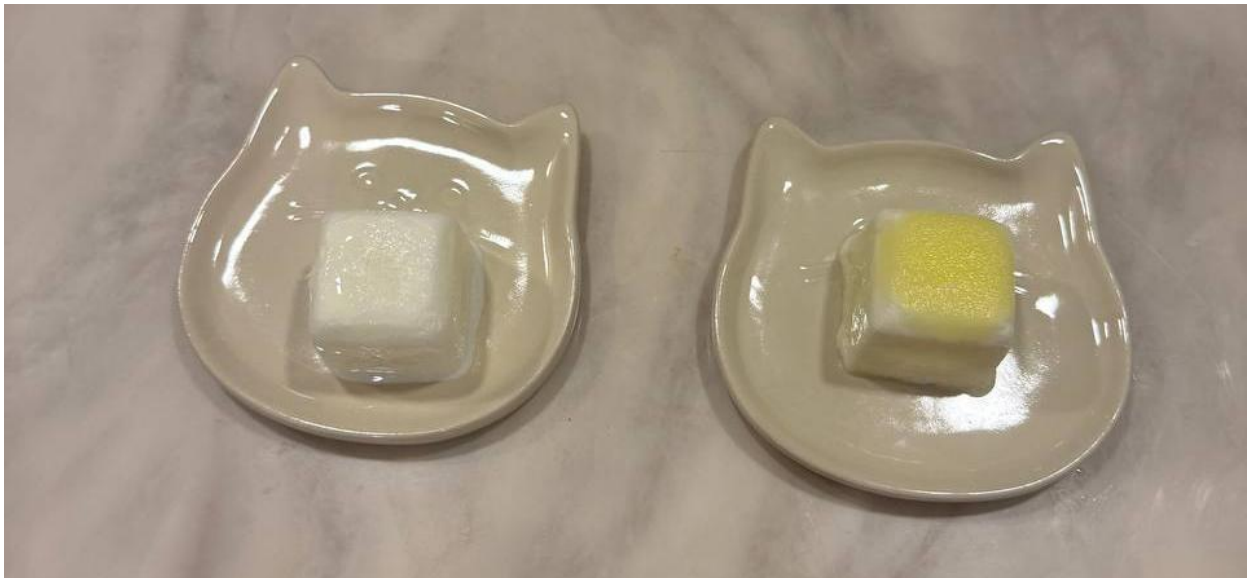

**Figure 1** Initial Samples Taken

**Limitations and problem faced**

With the limitation of camera, the video of the melting process is recorded in timelapse to save the storage. Every recording is recorded for 30 minutes, with frame rate of 3 per second, it is compressed to 20 seconds. This leads to less data samples can be analysed. Even with the smaller ice cream mould, the melting process is still long, hence it is not practical to record every video for a total of 30 min and send the huge data for analysis.

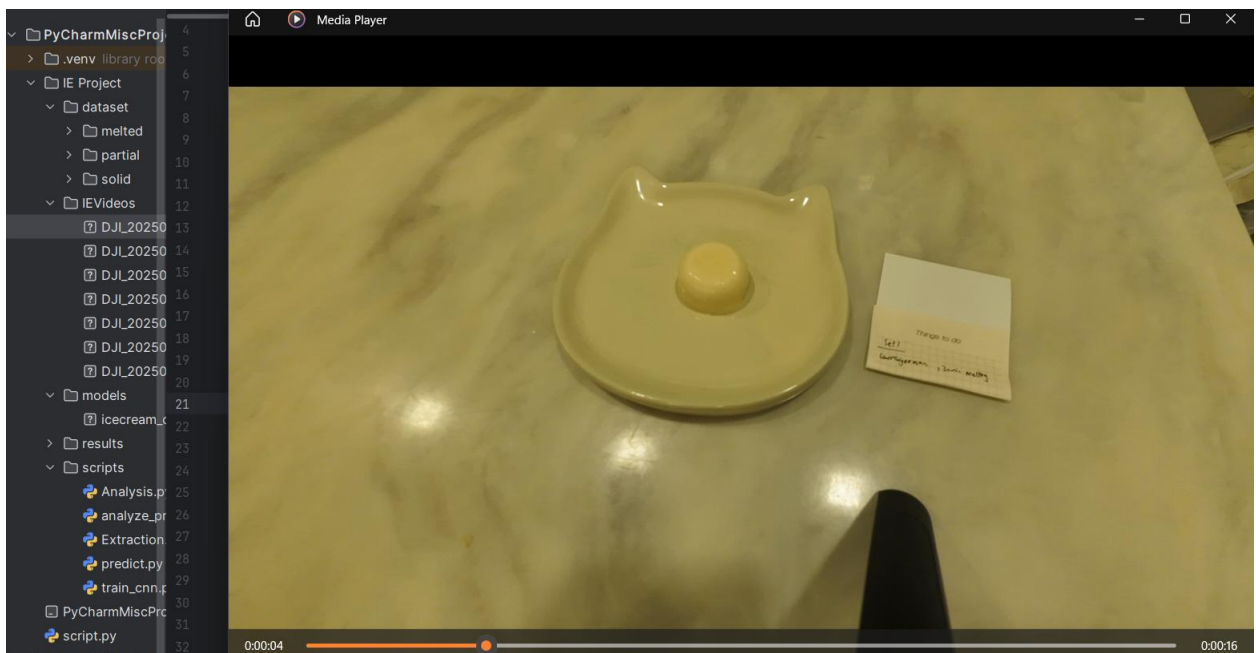

**Figure 2** Timelapse Video Extraction

Although there are many pre-trained CNN models, it is not always applicable and accurate for every project. In the initial stage of extraction, the images are sorted to the wrong files as the model could not visualize the state of ice cream accurately. Hence basic training of the CNN model is required. By manually arrange small number of images to the correct folder as shown in the picture below, the training begins. The CNN can visualize accurately the solid state apart from partial and melting. However, due to the strong structure of the samples with additives, the partial and melted image has not much of a big difference. As a result, when going through the dataset, the partial and melted image will still mix up. In-order to solve this problem, more data is to be recorded and upgrade of instruments with higher resolution and zoom is needed.

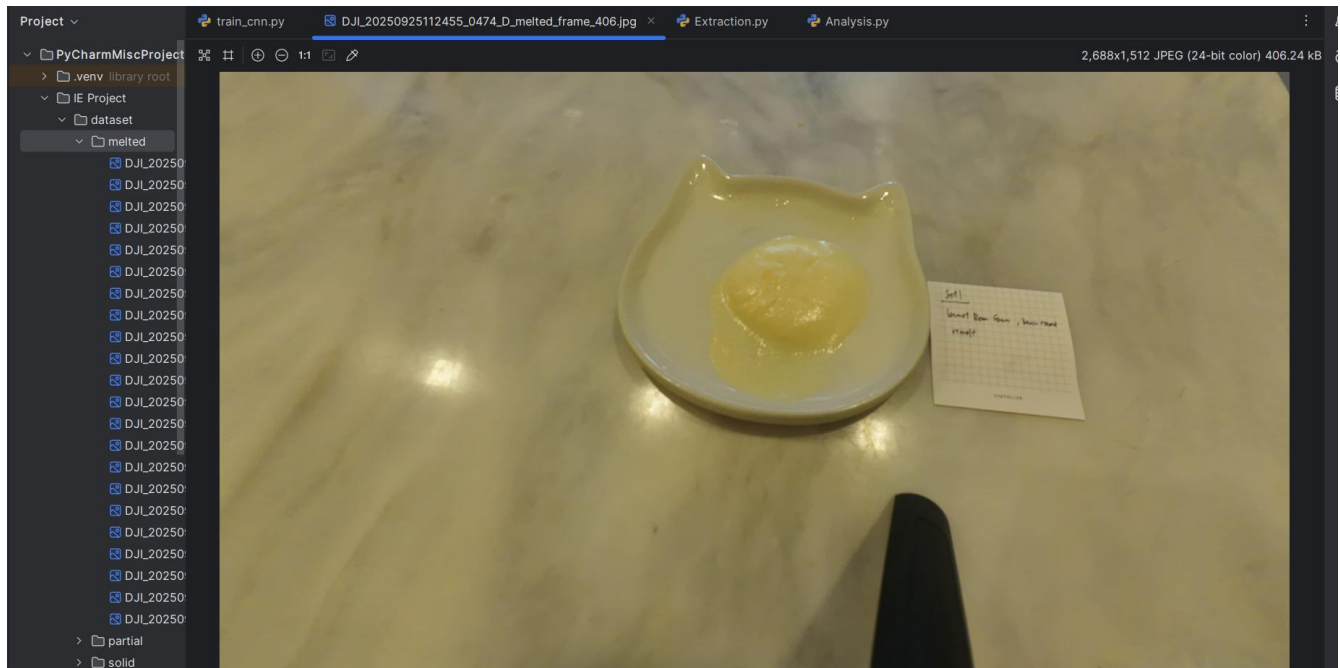

Figure 3 Manual Sorting Of Data For Training of CNN

Another limitation was the quantity of additives. As it is a small experiment, the sample size of each set is very small. Most of the recommended additive portion is between 0.01-0.3% each. Hence with small portions of ice cream mixture, the number of additives added to each sample that weights 40g are too little to be measured. Resulting in unprecise usage of additives. In large scale production or more scientific weighing scale can solve this problem.

## Results and findings

As mentioned in the above, the videos sample are extracted, and the images are classified according to its states.

```
Melting duration per sample (seconds):
predicted_class  melted  partial  solid
sample
DJI                 22       28      19
```

Figure 4 Sample Classification

In the figure 4 shown above, the state profile. It doesn't strongly resist melting (few remain solid), nor does it collapse immediately (not all melted). The fact is that the sample ice cream with additives can hold its shape and structure, the difference between partial and melted is not very big, hence it is difficult for the CNN to visualize. The figure below shows a completely melted sample with two additives. Its visual appearance remains almost unchanged compared to samples with no additives.

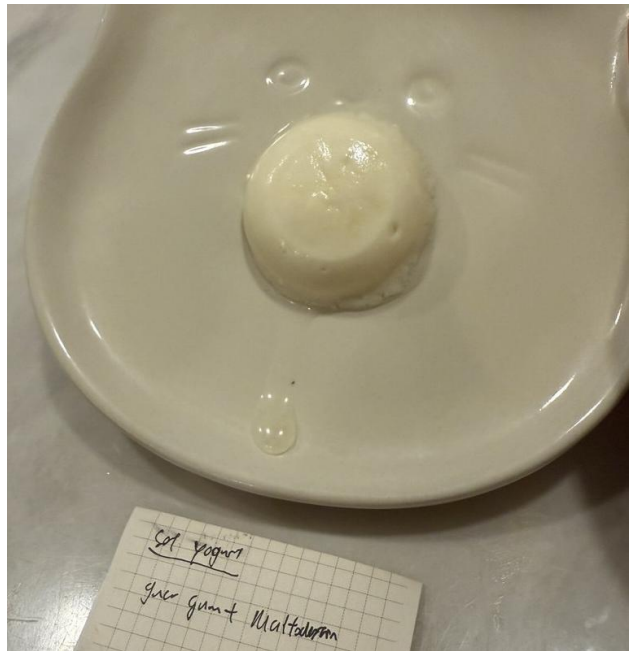
**Figure 5** Fully melted sample

In the figure 5 shows the melted yogurt sample, it has more liquid after melted but the ice cream remains in shape even tilted vertically, showing more stability in comparison to Set 1.

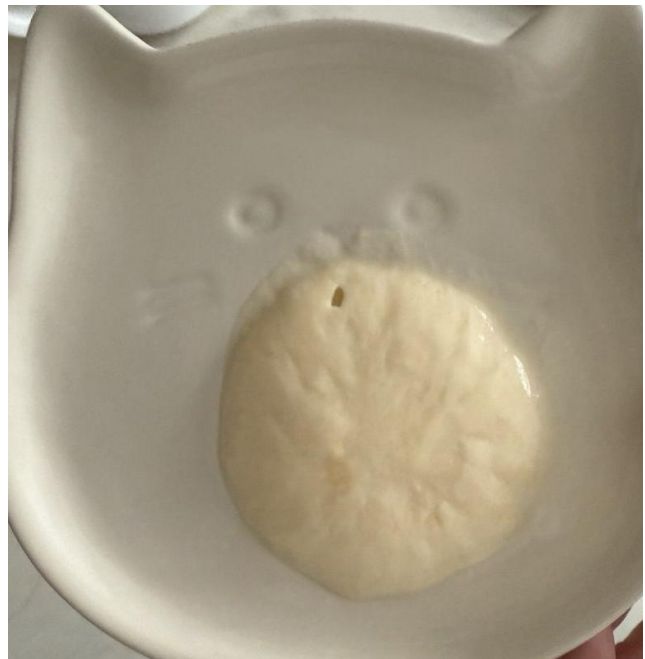
**Figure 6** Fully melted Set 1 sample after tilted

The sample after melted remains firm shapes, but after positioning the melted sample vertically, it starts to flow down slowly while shapes start to deform a bit.

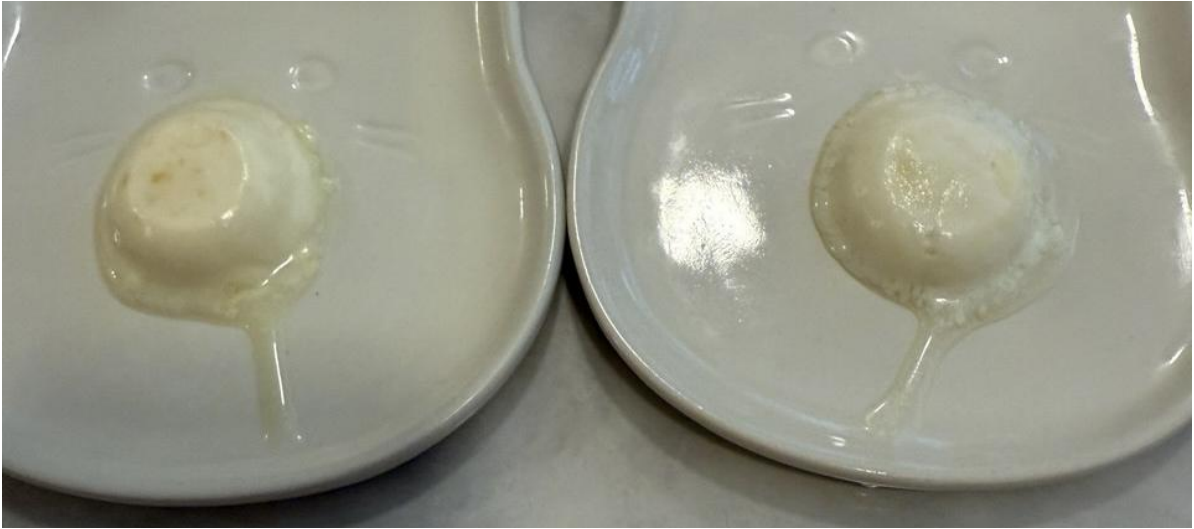

**Figure 7** Set 1 VS Set 2 (Both fully melted and has no additives)

Both set appeals to have more fluid as the additives help to lock up some of the water contents. However, by observing all the samples, the yogurt still has more fluid than the plain ice cream.

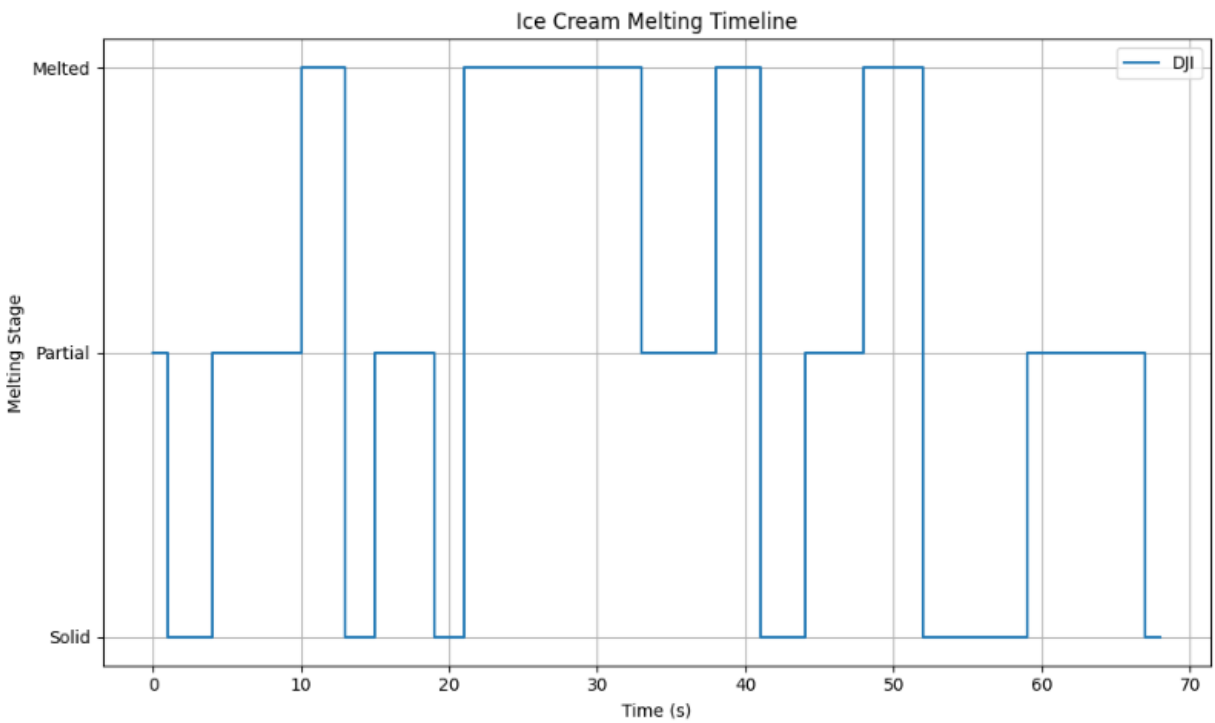

**Figure 8** Melting timeline plot

Based on the results shown in figure above, The plot is moving zigzagging, this happens because the model makes frame-by-frame predictions, and the frames are not smooth enough. In ideal cases, the progression from Solid to Partial to Melted can be observed without too much fluctuation. Overall, the plot shows that the samples partially melt within the first few seconds and become fully melted several times across the timeline. This happened due to errors in the model analyzation. The images analyze and conclude that the sample turn back to solid/partial state. But since CNN predictions have some uncertainty, this problem is not easy to solve in this project.

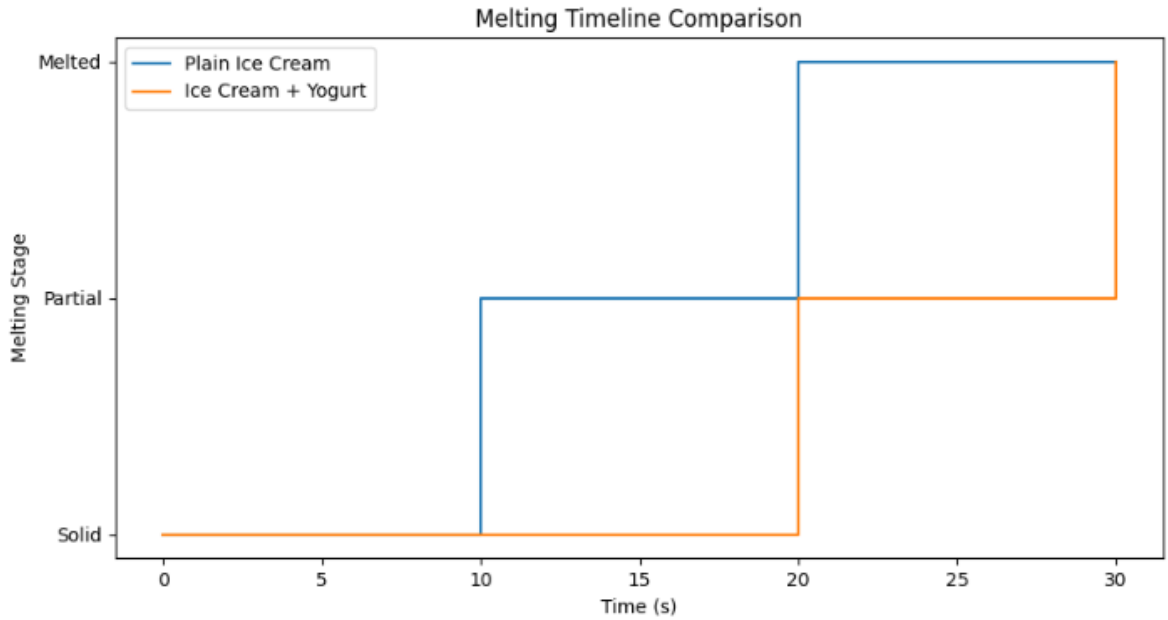

**Figure 9** Comparison plot

From the figure above, a simple comparison plot was seen, it shows clearly the plain ice cream melts faster. Its transitioning from Solid to Partial to Melted is around 20 min in real time. While yogurt ice cream shows higher stability, taking up to 30 min to fully melt. The addition of yogurt slows down the melting process at every stage. It delays transition from solid to partial and transition from partial to melted. However, this reading is not fully accurate, as the structure of yogurt ice cream is firmer, it is hard to gauge the actual state of the samples.

```
Epoch 6/10
4/4 ━━━━━━━━━━━━━━━━━━━━ 2s 452ms/step - accuracy: 0.7544 - loss: 0.5195 - val_accuracy: 0.5833 - val_loss: 0.9806
Epoch 7/10
4/4 ━━━━━━━━━━━━━━━━━━━━ 2s 454ms/step - accuracy: 0.7895 - loss: 0.4757 - val_accuracy: 0.5833 - val_loss: 0.9410
Epoch 8/10
4/4 ━━━━━━━━━━━━━━━━━━━━ 2s 449ms/step - accuracy: 0.7544 - loss: 0.4381 - val_accuracy: 0.6667 - val_loss: 1.0339
Epoch 9/10
1/4 ━━━━━━━━━━━━━━━━━━━━ 1s 481ms/step - accuracy: 0.8125 - loss: 0.3642
```

**Figure 10** Epoch Accuracy

From the figure above, as more samples are added, improvement can be seen. This improvement means the CNN model becomes increasingly reliable at classifying ice cream sample to solid, partial, or melted state as training progresses. The increase in training samples ensures that the network learns from a variety of visual differences, which increase it accuracy and reduces misclassification errors.

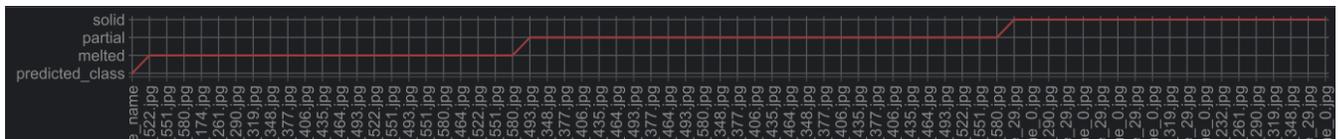

**Figure 11** Trained CNN splitting new data

The prediction of image after training the data. With more videos added, the model can take in the new data and classify the images according to its state, though by human observations, some of the partial and melted image is still classified wrongly. However, the solid-state classification is more accurate than before.

## Discussion

In this section, the results and finding of the experiments conducted are discussed, along with some highlights of limitations and improvement that can be done. With all the 18 samples going through the process of frozen, melting, refroze and re-melt. It is observed and recorded that Set 1 (plain ice cream) has much lesser water contents than Set 2 (yogurt ice cream). Set 1 after going through the first melting loss some moisture content to surround, hence at the second round of re-melt it becomes sturdier and stick well on the plate without dripping. At the same time, it remains a soft and foamy texture. While yogurt ice cream has too much water contents, it still drips and flow more easily than Set 1 and its texture remain almost the same as before. In consideration of production, transportation and consumer factors, the ice cream may experience melt and freeze more than once. Suitable container is important; at the same time the shape and flavour should undergoes least changes. Hence, I would recommend Set 1 more than Set 2 as Set 1 with lesser water contents are less like to flow and disform in the package. However, if the water contents in the yogurt can be removed in advance, I would like to propose Set 2 as the oats fibre plays a role in maintaining the shapes of the ice cream too and adds additional texture when consumed.

Looking into the price and results of the additives, I would recommend the usage of 2 additives. When using only one additive for example 0.4g of carrageenan (sample A) compared to two additives of 0.4g carrageenan and 0.4g of maltodextrin (sample B), the sample B have lesser water contents and does not affect the taste of ice cream. It also creates more stable structure than sample A. With the aid of other sample sets, it is observed that two additives perform better than one or four additives. When adding all the four additives in the sample, the taste and texture are not so great in comparison. Looking deeper in the combinations of additives, it is recommended to use guar gum and maltodextrin. They are both cheap and make a good combination in maintaining structures and has great taste and creaminess. Though carrageenan and guar gum make the cheapest pair and has better melt resistance, it does not show a significant difference to guar gum and maltodextrin pair.

In the PyCharm and CNN analysis, it is seen that the small difference between partial and melted is affecting the analysis and training. Making the results less accurate and hard to judge, the data plot produced are not refine and cannot be a reliable reference to use. Hence, more training data is needed, and manual adjustment is needed to increase the accuracy. Stronger and cleaner code can be used for processing more sets of different samples.

## Conclusion

This project investigated the melting behaviour of plain ice cream and yogurt ice cream with the addition of different stabilizers. The samples are collected and analysed through extraction of image and classification using a CNN model. The results showed that plain ice cream (Set 1) generally has lower water content and is less prone to dripping after repeated freeze and melt cycles, making it more suitable for production and transport. Yogurt-based ice cream (Set 2), while showing greater firmness in structure. However, it retains higher water content and flows more easily when melted, posing challenges for packaging and stability. Moreover, the cost of Set 1 will be much lesser than Set 2.

The comparison between additive combinations highlights that two additives perform better than single or multiple additive sets. Which also concludes that the set guar gum and maltodextrin set provided the best balance of creaminess, structure, and taste, while carrageenan and guar gum set showed slightly stronger melt resistance but less desirable texture. The choice made shows that the criterial of stabilizer selection depends on both sensory quality and melting resistance while retaining cheap price.

The CNN analysis demonstrated high potential for automated classification of melting states. With the accuracy improving significantly as more data and epochs were added, it is a desirable tool to work with. However, the small visual difference between partial and melted states reduced classification reliability, as the model often confused these two categories. Despite this limitation, the CNN was still effective and consistent at identifying solid states and provided useful insight in the melting progression.

This paper evaluates the results and data, looking into the enhancement of melt resistance ice cream product and choose the best option Set 1 recipe with guar gum and maltodextrin. It shows that the shape retained perfectly with the additives, production at a cheaper price and barely have any further deformation. Moreover, the texture become even firmer after excess liquid is lost in the melting process, making even creamier in the process.

**Data Availability**

All data generated or analysed during this study are included in this published article.